\documentclass[a2paper,11pt]{article}
\usepackage{spconf,amsmath,epsfig,amssymb,amsthm}
\usepackage{graphicx}
\usepackage{cite}
\usepackage{tabu}
\usepackage{multirow}
\usepackage{subfigure}
\usepackage[font=small]{caption}
\graphicspath{{./Figures/}}
\DeclareGraphicsExtensions{.pdf,.jpg,.png,.gif}
\usepackage{paralist, tabularx}
\DeclareMathOperator{\AP}{AP}
\DeclareMathOperator{\FP}{FP}

\title{Classification of remote sensing images using attribute profiles and feature profiles from different trees: a comparative study}

\name {Minh-Tan Pham$^1$, Erchan Aptoula$^2$, S\'ebastien Lef\`evre $^1$ \thanks{This work was supported by R\'egion Bretagne grant, the BAGEP Award of the Science Academy and the Tubitak Grant 115E857.}}

\address{
$^1$Universit\'e Bretagne Sud - IRISA, UMR 6074, 56000 Vannes, France\\
$^2$ Institute of Information Technologies, Gebze Technical University, 41400, Kocaeli, Turkey\\
\texttt{\{minh-tan.pham,sebastien.lefevre\}@irisa.fr,eaptoula@gtu.edu.tr}}

\begin{document}
%\ninept
%
\maketitle
\begin{abstract}
The motivation of this paper is to conduct a comparative study on remote sensing image classification using the morphological attribute profiles (APs) and feature profiles (FPs) generated from different types of tree structures. Over the past few years, APs have been among the most effective methods to model the image's spatial and contextual information. Recently, a novel extension of APs called FPs has been proposed by replacing pixel gray-levels with some statistical and geometrical features when forming the output profiles. FPs have been proved to be more efficient than the standard APs when generated from component trees (max-tree and min-tree). In this work, we investigate their performance on the inclusion tree (tree of shapes) and partition trees (alpha tree and omega tree). Experimental results from both panchromatic and hyperspectral images again confirm the efficiency of FPs compared to APs.
\end{abstract}
%
%\setlength{\belowdisplayskip}{1.5pt} \setlength{\belowdisplayshortskip}{1.5pt}
%\setlength{\abovedisplayskip}{1.5pt} \setlength{\abovedisplayshortskip}{1.5pt}
%========================================================================
\section{Introduction}
\label{sec:intro}
Since their first introduction to remote sensing field in early 2010's, morphological attribute profiles (APs) \cite{dalla2010morphological} have been widely used thanks to their powerful multilevel modeling of spatial information from the image content and their efficient implementation via tree structures. By well preserving important spatial properties of regions and objects such as contours, shape, compactness, etc., APs characterize effectively the contextual information of the observed scene, hence become relevant for remote sensing image analysis, especially for classification task. In the past few years, a great number of research studies have been devoted to improve their classification performance. Some have focused on modifying the AP construction framework, while others have attempted to post-process the output profiles to increase their description capacity (i.e. see a recent survey in \cite{pham2018recent}). 

It is worth noting that in all standard AP-based methods, the output profiles are a set of filtered images obtained by the tree-based attribute filtering process. Hence, they are still the gray values of the connected components (CCs) w.r.t the nodes of the filtered trees. In order to provide better characterization of the CCs, the feature profiles (FPs) \cite{pham2017feature} have been recently developed. Instead of reconstructing the filtered image using pixel gray values from the pruned tree as in APs, FPs extract some statistical features (i.e. mean, standard deviation, entropy, etc.), together with some geometric and shape information (i.e. area, elongation, diagonal length of bounding box, etc.), hence involving more complete information. The superior performance of FPs has been validated compared to the standard APs, i.e. constructed from component trees (max-tree and min-tree). In the present work, our objective is to investigate this behavior on other tree structures, in particular the inclusion tree (tree of shapes) and partition trees (alpha tree and omega tree). This comparative study helps to confirm the effectiveness of this promising extension.

In the remainder of this paper, the backgrounds of APs and FPs are revised in Section \ref{sec:method}. In Section \ref{sec:result}, we describe the experimental setup and provide comparative results of supervised classification conducted on both panchromatic Reykjavik image and hyperspectral Pavia University data.  Section \ref{sec:conclusion} finally concludes the paper and discusses some further work.

\begin{figure*}[]
\centerline{\begin{minipage}[b]{0.95\linewidth}
  \centering
  \includegraphics[width=160mm]{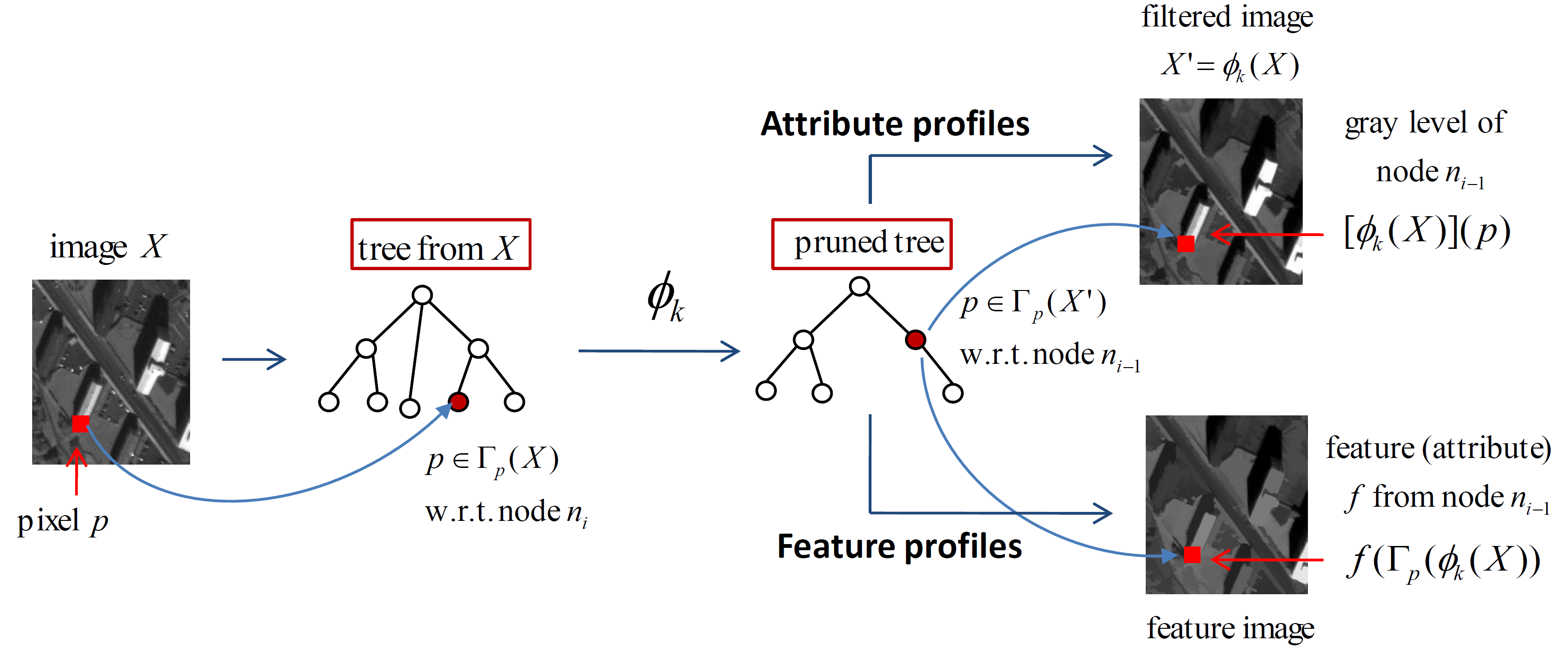}            
\end{minipage}}
\caption{Generation of the APs and FPs from a grayscale image $X$ w.r.t. the attribute filtering $\phi_k$. Here $\Gamma_p$ denotes the connected component (CC) containing $p$; $X' = \phi_k(X)$ is the filtered image; and $f$ is the feature (attribute) to be extracted.}
\label{fig:AP_FP}
\end{figure*}

\section{Morphological attribute profiles and feature profiles}
\label{sec:method}
\subsection{Morphological Attribute Profiles}
APs are multilevel image description tools obtained by successively applying a set of morphological attribute filters (AFs)  \cite{dalla2010morphological}. Unlike usual image filtering operators which are directly performed on pixel level, AFs work on CC level based on the concept of image connectivity. In other words, an AF is a filtering operator applied on the tree's node level with regard to a specific attribute characterizing the size, shape, or other properties of objects present in the image. 
%To this end, the generation of standard APs from an image can be summarized as a four-step process: 1) construct the max-tree (and the min-tree) from the image; 2) compute some relevant attributes (area, moment of inertia, standard deviation, etc.) from each node of the tree; 3) filter the tree by keeping/removing nodes according to their attribute values; 4) reconstruct the image from the filtered tree. Steps 3) and 4) can be done for different attributes (with different threshold values) to finally produce a set of filtered images (by stacking them) to form APs.

Given a grayscale image $X:E \rightarrow \mathbb{Z}, E \subseteq \mathbb{Z}^2$, the standard generation of APs on X is achieved by applying a sequence of AFs based on a min-tree (attribute thickening operators $\{\phi_k\}_{k=1}^K$) and a max-tree (i.e. attribute thinning operators $\{\gamma_k\}_{k=1}^K$). The AP descriptor of each pixel $p$ in the definition domain of $X$ is written:
\begin{equation}
\begin{split}
\AP(p) =
\Big\{ \big[\phi_K(X)\big](p),\ldots,\big[\phi_1(X)\big](p),X(p),  \\
		\big[\gamma_1(X)\big](p),\ldots,\big[\gamma_K(X)\big](p) \Big\}.
\end{split}
\label{eq:ap_p}
\end{equation}
%\begin{equation}
%\begin{split}
%\AP(X) = \Big\{\phi_K(X),\phi_{K-1}(X),\ldots,\phi_1(X),X, \\
%		 \gamma_1(X),\ldots,\gamma_{K-1}(X),\gamma_K(X) \Big\},
%\end{split}
%\label{eq:ap}
%\end{equation}
where $\phi_k(X)$ is the filtered image obtained by applying the attribute thickening $\phi$ with regard to the threshold $k$. Similar explanation is made for $\gamma_k(X)$. As observed, the feature dimension of $\AP(p)$ is $(2K+1)$.

%the resulted $\AP(X)$ is a stack of $(2K+1)$ images including the original image, $K$ filtered images from the thickening profiles and the other $K$ from the thinning profiles. For more details about this AP computation, readers are referred to papers \cite{dalla2010morphological}. 

\subsection{Feature Profiles}
In order to better characterize the region/object enclosed by the CC (i.e. which corresponds to a filtered tree's node), node features are extracted instead of the node's gray level in the recently proposed FPs \cite{pham2017feature} . Figure \ref{fig:AP_FP} provides an overview of how the generation of FPs differs from the standard AP technique. In fact, after obtaining the pruned tree by an attribute filtering $\phi_k$, instead of reconstructing the filtered image using the nodes' gray levels, different features are outputed to form FPs. 

Specifically, for each pixel $p$, AP of $p$ obtained by an arbitrary AF $\phi_k$ is the gray value $X'(p)$, where $X'=\phi_k(X)$ is the image reconstructed from the filtered tree (cf. Eq \eqref{eq:ap_p}). Now, let $\Gamma_p(X)$ be the CC of $X$ containing $p$ and let $f$ be a feature or an attribute, i.e. a function admitting a CC and outputting a real value, to be extracted. The FP of $p$ will be $f[\Gamma_p(X')]$. More formally, the generation of FPs w.r.t the feature $f$ based on min-tree and max-tree are defined as follows:
\begin{equation}
\begin{split}
\FP_f(p) = \Big\{ f\big[\Gamma_p(\phi_K(X))\big],\ldots,f\big[\Gamma_p(\phi_1(X))\big], \\
X(p), f\big[\Gamma_p(\gamma_1(X))\big],\ldots,f\big[\Gamma_p(\gamma_K(X))\big] \Big\}.
\end{split}
\label{eq:fp_p}
\end{equation}

Unlike in AP technique where only one profile is produced from a pruned tree, several features can be simultaneously extracted and stacked to form the final FP:
\begin{equation}
\FP_{f_1+\ldots+f_n}(p) = \Big[ \FP_{f_1}(p), \ldots, \FP_{f_n}(p) \Big].
\label{eq:fp_multi}
\end{equation}

\subsection{APs and FPs on other trees}
Instead of calculating the APs based on both max-tree and min-tree image representation as in the original work \cite{dalla2010morphological}, other implementations have been proposed using the inclusion tree (tree of shapes) \cite{dalla2011self} (i.e. self-dual APs or $sd$-APs) as well as the partition trees such as $\alpha$-tree and $\omega$-tree to produce $\alpha$-APs, $\omega$-APs, respectively \cite{bosilj2017attribute}. The advantage of  these trees is that their self-dual property enables the attribute filtering to simultaneously access and model both dark and bright
regions from the image. Thus, only one tree per image is required instead of both max-tree and min-tree as in \cite{dalla2010morphological}. The dimension of their APs is therefore reduced by half. Moreover, partition trees offer the possibility to work on multivariate images only using a single tree, which is not trivial by exploiting component or inclusion trees \cite{bosilj2017attribute}.

In this work, those different tree structures are also adopted for the implementation of FPs. The so-called self-dual FPs ($sd$-FPs), $\alpha$-FPs, $\omega$-FPs, respectively, are defined by generating FPs from the tree of shapes, $\alpha$-tree and $\omega$-tree. The motivation is now to investigate the performance of these FP alternatives compared to their AP counterparts.

\section{Experimental study}
\label{sec:result}
Supervised classification was performed using both panchromatic and hyperspectral remote sensing images in our experiments. We first introduce the two data sets and the experimental setup. Then, the comparative evaluation of classification results achieved by APs and FPs based on different trees is provided.

\subsection{Data sets and experimental setup}
\begin{figure}[!ht]
\centerline{\begin{minipage}[b]{0.9\linewidth}
  \centering
  \includegraphics[width=24mm]{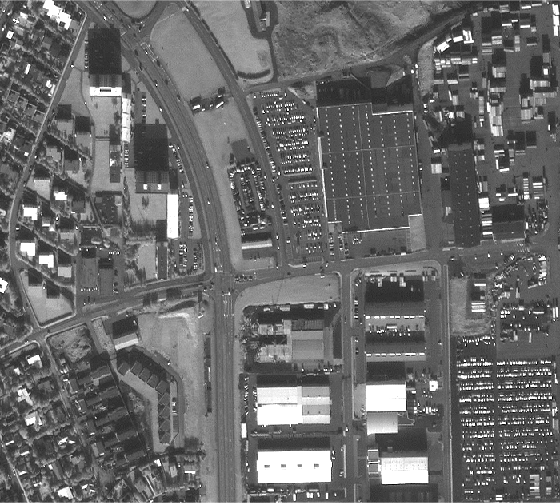}
  \hfill
  \includegraphics[width=24mm]{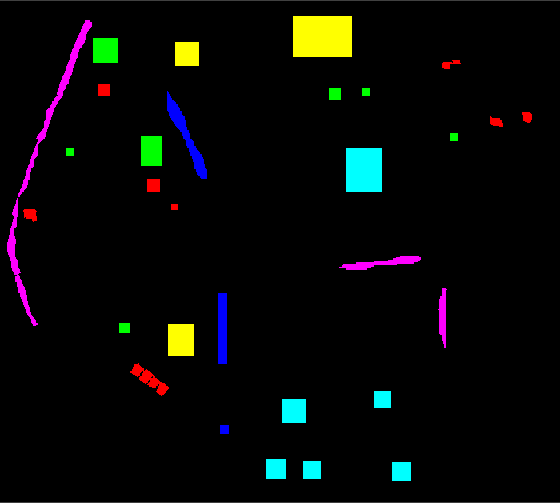}
  \hfill
  \includegraphics[width=24mm]{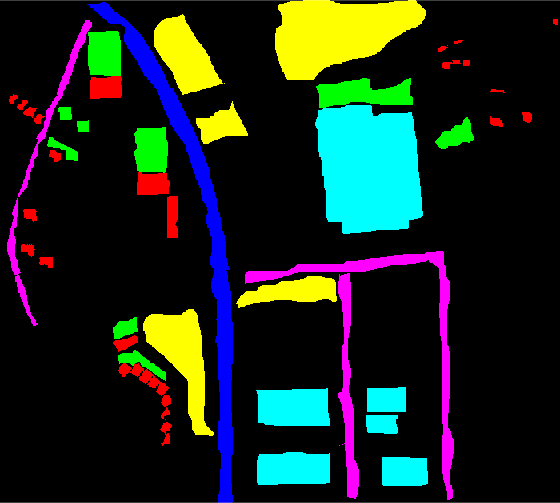} 
  \vfill
%  \vspace{1mm}
  \includegraphics[trim=35mm 90mm 30mm 72mm, clip, width=80mm]{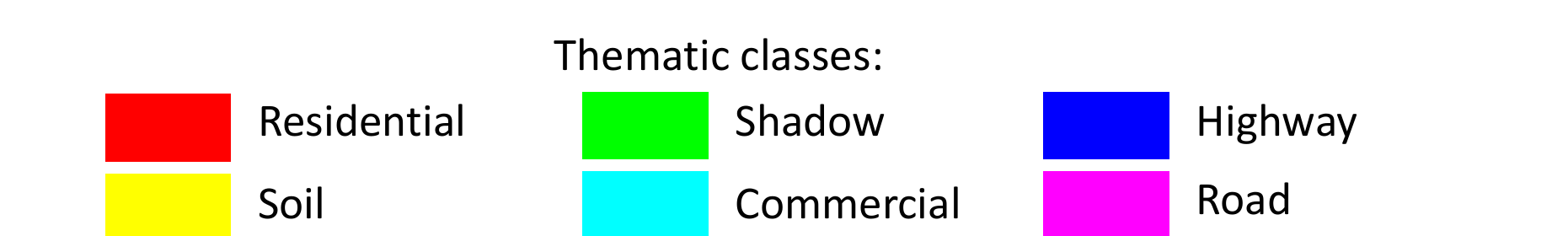}
  \vfill
  \footnotesize{\textbf{(a)}}             
\end{minipage}}
\vfill
%\vspace{2mm}
\centerline{\begin{minipage}[b]{0.9\linewidth}
  \centering
  \includegraphics[width=24mm]{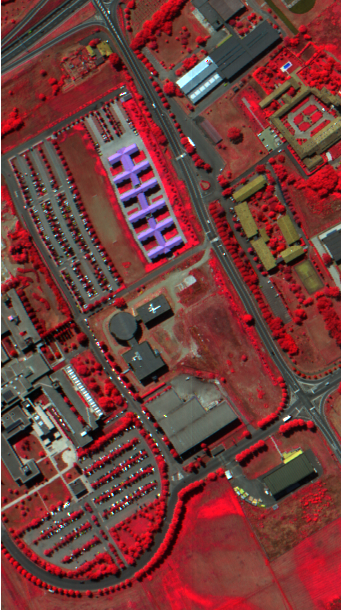}
  \hfill
  \includegraphics[width=24mm]{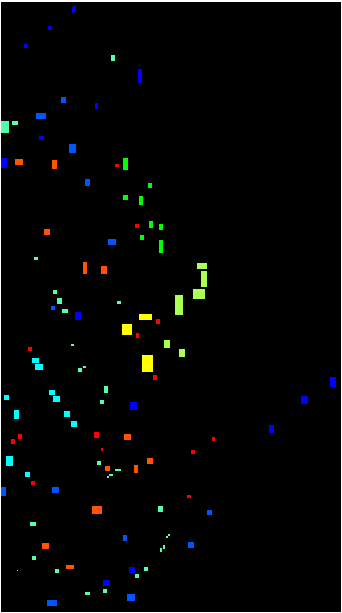}
  \hfill
  \includegraphics[width=24mm]{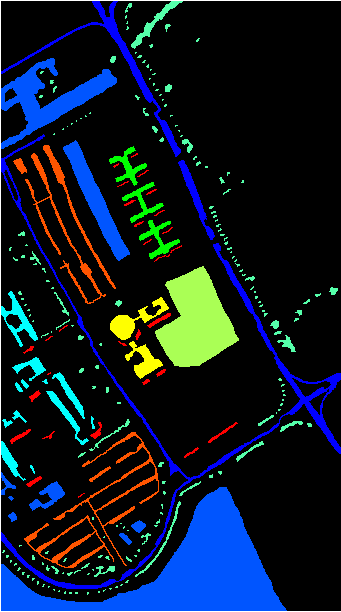} 
  \vfill
%  \vspace{1mm}
  \includegraphics[trim=40mm 78mm 30mm 72mm, clip, width=80mm]{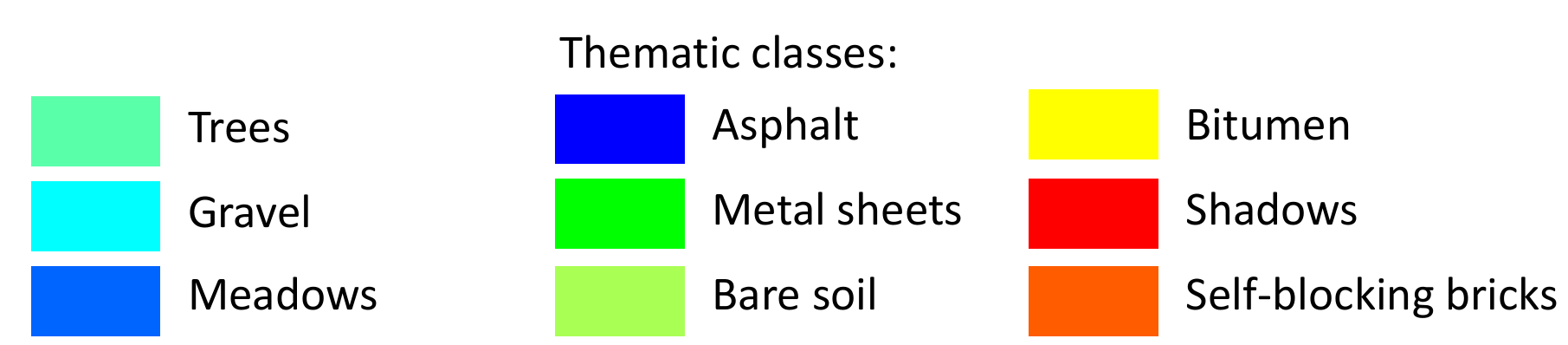}
  \vfill
  \footnotesize{\textbf{(b)}}             
\end{minipage}}
\caption{Two experimental data: \textbf{(a)} The Reykjavik data (left to right: panchromatic, training samples and test samples); \textbf{(b)} The Pavia University data (left to right: false-color image made by bands 31-56-102, training samples and test samples).}
\label{fig:dataset}
\end{figure}

Two image data used in our experiments are shown in Figure \ref{fig:dataset}. The first one is a panchromatic Reykjavik image of size $628 \times 700$ pixels acquired by the IKONOS Earth imaging satellite with 1-m resolution. This image consists of six thematic classes including residential, soil, shadow, commercial, highway and road. For classification task, 22741 training and 98726 test pixel samples were considered as in the figure (a). The second one is the hyperspectral Pavia University image acquired by the ROSIS airborne sensor with 1.3-m spatial resolution. The image consists of $610 \times 340$ pixels with 103 spectral bands (from 0.43 to 0.86 $\mu$m) and covers nine thematic classes: trees, asphalt, bitumen, gravel, metal sheets, shadows, meadows, self-blocking bricks and bare soil. From this image, 3921 training and 42776 test samples were split for classification experiments.

For attribute filtering, we exploited two attributes including the \emph{area} and the \emph{moment of inertia}. Then, both statistical (standard deviation) and geometrical (area) features were extracted to form FPs. We note that other statistical or geometrical features could be extracted as well. Here, standard deviation and area were selected thanks to their stable performance from lots of our experiments. The attribute threshold values were set as in \cite{pham2017feature,pham2017classification,pham2017local}. For the hyperspectral Pavia image, the first four PCA components were exploited as in most of related work.
To perform supervised classification, the output APs and FPs generated from different tree structures (component, inclusion and partition trees) were fed into the random forest classifier. The number of trees was set to 100. Standard implementation as well as equivalent parameter configuration were performed to ensure a fair comparison. Finally, to report the classification performance, the overall accuracy (OA) and the kappa coefficient ($\kappa$) were considered.

\subsection{Comparative results and discussion}
Tables \ref{tab:res_reykjavik} and \ref{tab:res_pavia} report the classification results of the Reykjavik and the Pavia data, respectively, yielded by APs and FPs over different trees. One general remark is that these methods are quite sensitive to the types of tree and attribute to use but in most of the cases, FPs provided better results than APs. The best performance for the Reykjavik image was achieved by FPs from $\alpha$-tree with an OA equal to $88\%$ and for the Pavia image, it was obtained by FPs from component trees with $96.5\%$.
\begin{table}[!ht]
	\centering
	\resizebox{0.43\textwidth}{!}{%
	{\renewcommand{\arraystretch}{1.0}
	\begin{tabu}{l| r r| r r| r r}
	\hline
	\multirow{2}{*}{\textbf{Method}} & \multicolumn{2}{c|}{\textbf{Area}} & \multicolumn{2}{c|}{\textbf{Moment}} & \multicolumn{2}{c}{\textbf{Both}}\\
	\cline{2-7}
	 & OA & $\kappa$ & OA & $\kappa$ & OA & $\kappa$ \\
	 \hline
	$\AP$ & 81.5 & 0.767 &72.5 & 0.656 & 82.5 & 0.779\\
%	$\FP_{\mu+\sigma}$ & 85.0 & 0.809 &76.7 & 0.708 & 84.7 & 0.806\\
	$\FP$ & 85.3 & 0.813 &79.4 & 0.741 & 86.1 & 0.823\\
%	& \multicolumn{2}{c|}{\textbf{+3.8}} & \multicolumn{2}{c|}{\textbf{+6.9}} & \multicolumn{2}{c}{\textbf{+3.6}} \\ 
	%$\FP_{\mu+\sigma+a}$ \\
	\hline
	$sd$-$\AP$ & 82.5 & 0.780 & 62.5 & 0.530 & 82.3 & 0.778\\
	%$sd$-$\FP_{\mu+\sigma}$ & 83.1 & 0.787 &78.9 & 0.735 & 83.7 & 0.795\\	
	$sd$-$\FP$ & 83.1 & 0.786 &80.3 & 0.751 & 84.9 & 0.808\\
	%$sd$-$\FP_{\mu+\sigma+a}$ \\
	\hline
	$\alpha$-$\AP$ & 77.7 & 0.716 & 71.6 & 0.636 & 76.9 & 0.707\\
	%$\alpha$-$\FP_{\mu+\sigma}$ & 85.2 & 0.813 & 86.3 & 0.825 & 87.5 & 0.841\\
	$\alpha$-$\FP$ & 85.3 & 0.814 & \textbf{86.2} & \textbf{0.823} & \textbf{88.0} & \textbf{0.848}\\
	%$\alpha$-$\FP_{\mu+\sigma+a}$ \\
	\hline
	$\omega$-$\AP$ & 77.3 & 0.709 & 75.6 & 0.691 & 77.0 & 0.708\\
	%$\omega$-$\FP_{\mu+\sigma}$ & 84.0 & 0.797 & 88.4 & 0.852 & 88.1 & 0.849\\
	$\omega$-$\FP$ & \textbf{85.7} & \textbf{0.815} & 86.0 & 0.821 & 86.2 & 0.824\\
	%$\omega$-$\FP_{\mu+\sigma+a}$ \\
	\hline
	\end{tabu} 
	}}
	\caption{Comparison of classification performance on the Reykjavik data yielded by APs and FPs from different trees.}
	\label{tab:res_reykjavik}
	%}
\end{table}

Another important remark is that when using the moment attribute, FPs could consistently and significantly improve  the classification accuracy compared to APs, for both data and for all tree kinds as well. When filtering by the area, FPs outperformed APs from all trees for Reykjavik. For Pavia, FPs are more suitable when using component trees and $\alpha$-tree while the other trees are in favor of APs. To summarize, the comparative results from both tables have confirmed the effectiveness and good potential of FPs compared to APs.
\begin{table}[!ht]
	\centering
	\resizebox{0.43\textwidth}{!}{%
	{\renewcommand{\arraystretch}{1.0}
	\begin{tabu}{l| r r| r r| r r}
	\hline
	\multirow{2}{*}{\textbf{Method}} & \multicolumn{2}{c|}{\textbf{Area}} & \multicolumn{2}{c|}{\textbf{Moment}} & \multicolumn{2}{c}{\textbf{Both}}\\
	\cline{2-7}
	 & OA & $\kappa$ & OA & $\kappa$ & OA & $\kappa$ \\
	 \hline
	$\AP$ & 93.1 & 0.908 & 78.8 & 0.732 & 93.3 & 0.912\\
	%$\FP_{\mu+\sigma}$ & 85.0 & 0.809 &76.7 & 0.708 & 84.7 & 0.806\\
	$\FP$ & \textbf{96.5} & \textbf{0.954} & 84.7 & 0.804 & \textbf{96.4} & \textbf{0.953}\\
	%$\FP_{\mu+\sigma+a}$ \\
	\hline
	$sd$-$\AP$ & 91.8 & 0.892 & 75.7 & 0.694 & 92.5 & 0.901\\
	%$sd$-$\FP_{\mu+\sigma}$ & 83.1 & 0.787 &78.9 & 0.735 & 83.7 & 0.795\\	
	$sd$-$\FP$ & 91.4 & 0.888 & 82.7 & 0.778 & 91.7 & 0.891\\
	%$sd$-$\FP_{\mu+\sigma+a}$ \\
	\hline
	$\alpha$-$\AP$ & 90.7 & 0.879 & 88.6 & 0.853 & 92.9 & 0.907\\
	%$\alpha$-$\FP_{\mu+\sigma}$ & 85.2 & 0.813 & 86.3 & 0.825 & 87.5 & 0.841\\
	$\alpha$-$\FP$ & 94.7 & 0.923 & 94.9 & 0.926 & 95.3 & 0.930\\
	%$\alpha$-$\FP_{\mu+\sigma+a}$ \\
	\hline
	$\omega$-$\AP$ & 95.5 & 0.941 & 88.4 & 0.851 & 94.8 & 0.932\\
	%$\omega$-$\FP_{\mu+\sigma}$ & 84.0 & 0.797 & 88.4 & 0.852 & 88.1 & 0.849\\
	$\omega$-$\FP$ & 91.5 & 0.873 & \textbf{95.4} & \textbf{0.933} & 92.6 & 0.889\\
	%$\omega$-$\FP_{\mu+\sigma+a}$ \\
	\hline
	\end{tabu} 
	}}
	\caption{Comparison of classification performance on the Pavia data yielded by APs and FPs from different trees.}
	\label{tab:res_pavia}
	%}
\end{table}
%-------------------------------------------------------------------------------------------
\section{Conclusion}
\label{sec:conclusion}
We have revisited the principles of APs \cite{dalla2010morphological} and FPs \cite{pham2017feature} with the aim to conduct a comparative study of their performance on remote sensing image classification. Our experiments have taken into account various tree structures for their generation including the component, inclusion and partition trees. Experimental results on both panchromatic and hyperspectral images have confirmed the superior performance of FPs and thus revealed a high potential of this extension. Future work may focus on investigating FPs on other kinds of remote sensing data or combining them with deep neural networks.

%%morphological attribute profiles in the context of remote
%sensing image classification. Experimental study
%on one panchromatic and one hyperspectral image has been
%performed to provide a general evaluation of different methods
%compared to the original framework. This paper may serve as
%an overview of AP recent advances to readers as well as a
%guidance to researchers working on this framework and its
%alternatives within their work. We believe the exploitation and
%adaptation of APs in remote sensing imagery still remains an
%open research topic for on-going as well as future work.

\begin{small}
\section{Acknowledgement}
The authors would like to thank Prof. Jon Atli Benediktsson and Prof. Paolo Gamba for making available the Reykjavik image and the Pavia University data.
\end{small}
%--------------------------------------------------------------
\begin{footnotesize}
\bibliographystyle{ieeetr}
\bibliography{RefAbrv,RefAPs}
\end{footnotesize}
\end{document}